# Online influence, offline violence: Language Use on YouTube surrounding the 'Unite the Right' rally

Isabelle van der Vegt[1]    Maximilian Mozes[1]    Paul Gill[1]    Bennett Kleinberg[1,2]

[1]Department of Security and Crime Science, University College London
[2]Dawes Centre for Future Crime, University College London
Correspondence: isabelle.vandervegt@ucl.ac.uk

*Abstract*— The media frequently describes the 2017 Charlottesville 'Unite the Right' rally as a turning point for the alt-right and white supremacist movements. Social movement theory suggests that the media attention and public discourse concerning the rally may have influenced the alt-right, but this has yet to be empirically tested. The current study investigates whether there are differences in language use between 7,142 alt-right and progressive YouTube channels, in addition to measuring possible changes as a result of the rally. To do so, we create structural topic models and measure bigram proportions in video transcripts, spanning eight weeks before to eight weeks after the rally. We observe differences in topics between the two groups, with the 'alternative influencers' for example discussing topics related to race and free speech to an increasing and larger extent than progressive channels. We also observe structural breakpoints in the use of bigrams at the time of the rally, suggesting there are changes in language use within the two groups as a result of the rally. While most changes relate to mentions of the rally itself, the alternative group also shows an increase in promotion of their YouTube channels. Results are discussed in light of social movement theory, followed by a discussion of potential implications for understanding the alt-right and their language use on YouTube.

I. INTRODUCTION

On 11 and 12 August 2017, dozens of alt-right, white supremacist and neo-Nazi individuals descended on Charlottesville, Virginia. The event, known as the 'Unite the Right' rally, turned fatal on the second day when a white supremacist deliberately drove into a crowd of counter-protestors, resulting in the death of one person and leaving several others injured [1, 2]. In recent years, the rise of the alt-right has been accompanied by several other acts of violence and terror attacks motivated by white supremacist ideologies, with 18 out of 34 extremist-related deaths in 2017 attributed to this group [3]. In 2018, all 50 extremist murders in the United States were linked to right-wing extremism [4].

At the same time, alt-right ideologies have become widespread online. Their content is easily accessible through social media platforms, and ideas are amplified on websites such as 4chan [5] and Gab [6]. YouTube, in particular, has been described as a breeding ground for the alt-right [7, 8]. Online political influencers adopt strategies of mainstream popular Youtubers to gain popularity, engaging in tactics for search engine optimisation and cultivating a relatable 'underdog' image [8]. This paper examines language use on YouTube, in a unique dataset of 'alternative influencers' and progressive channels. We discuss differences between and within these groups, in a timeframe shortly before and after the Charlottesville rally.

II. BACKGROUND

*A. The alt-right and Charlottesville*

The alt-right is not defined by a central organisation [9], nor does it 'offer a coherent or well-developed set of policy proposals' [10]. Instead, it has been referred to as a 'mix of rightist online phenomena' [11] with white identity at its core [12]. The alt-right is variously characterised as anti-political correctness, anti-immigration, anti-Semitist, and anti-feminist [12], ideologies which are commonly spread online through irony and dark humour. Scholars have begun to study the alt-right as a social movement, following the definition of 'a cluster of performances organised around a set of grievances or claims' [9, 13]. It has been argued that the alt-right, mainly through online activity, engages in promoting a shared identity, fostering commitment to a common cause, and proclaiming the 'worthiness, unity, and size' of its movement [9].

After the Charlottesville rally, various media outlets declared that 'white nationalists are winning' [14] and 'the genie is out of the bottle' [1]. In addition, President Trump stated that there was 'blame on both sides' [15], which prompted the suggestion that his claims 'reinvigorated' the alt-right movement [15]. In the aftermath of the rally, various reports also noted that white nationalists have entered mainstream conversation [1, 16] and some say they were aided in doing so by the Trump administration [16].

*B. Social movement theory*

Elements of social movement theory propose that people engage in social identity performance, which refers to behaviour that serves to express the norms of the social group one aims to belong to [17, 18]. Such behaviour includes affirming ones social identity, conforming to a social movement, strengthening ones identity, or mobilising others [17]. Within the context of the alt-right, social identity performance may, for example, include using community-specific language [5] or memes online (e.g., Pepe the Frog, a popular internet meme appropriated by the alt-right [5, 10]), or to publicly adopt symbols related to white nationalism.

Research on the effect of media coverage and public discourse on social movements might explain the potential effect of the rally on the alt-right [19, 20]. For example, research on right wing violence in Germany suggests that positive and negative reactions from public figures to violent events may help to lend prominence to the movement [20]. That is, even if one aims to condemn a violent movement's message, the message is (at least partially) reproduced [20]. By studying newspaper sources, this line of research suggested that discursive opportunities, summarised as public visibility, resonance, and legitimacy affected the behaviour of right-wing movements, measured in terms of violent events against different target groups [20]. Public visibility refers to the number of outlets reporting on the movement and the prominence of the movement's message within those outlets [20]. Resonance is defined as the (positive or negative) reaction from public figures to the movement's message as well as the associated ripple effect in the media [20]. Legitimacy involves the general public's support of a message [20]. Similar discursive opportunities were also studied in relation to the rise in popularity of right-wing populist Pim Fortuyn in the Netherlands [19]. In a similar vein, visibility (e.g. the extensive media coverage), resonance (e.g. responses to the rally from President Trump and other politicians), and legitimacy (e.g. subsequent protests and vigils denouncing the rally [21]) can be observed in the context of the alt-right and Charlottesville rally.

The effect of discursive opportunities has yet to be examined for the specific case of the alt-right and the Charlottesville rally. If indeed the visibility of the alt-right increased following the rally, the message of the movement resonated in the media and public discourse, and the alt-right gained legitimacy through acknowledgement from opponents and the general public, we may expect to see changes in behaviour within the movement. Within the context of social identity performance, one may expect to see strengthened social identity consolidation within the alt-right movement as a result of the rally, President Trump's comments, and the media coverage of the rally. After the rally, we might expect increased expression of norms from the alt-right movement, for example in the form of stronger endorsement or more extreme expressions of in-group ideology. As has been raised previously, such behaviour may serve to further strengthen the movement or mobilise others to join. Empirical examinations of these possibilities are thus far lacking. Moreover, reactions to the rally have yet to be comprehensively examined for YouTube, seen as one of the hotbeds of alt-right online activity [8]. One way in which reactions to an event within a movement can be measured is by modelling language use surrounding an event of interest. In the following section, we discuss previous efforts at doing so.

*C. Linguistic effects of exogenous events*

The effect of extremism-related events on social media behaviour has been previously examined in a few cases. In a qualitative study of Twitter accounts of two alt-right and one far-left organisation in the six weeks leading up to the Charlottesville rally, it was observed that the two sides frequently targeted each other, framing the opposing group as the enemy [22]. Manual examination of the tweets showed that the alt-right accounts frequently referred to 'the left' and 'liberals' as unpatriotic and communist. At the same time, the far-left accounts dubbed the alt-right 'suit and tie Nazis'. Furthermore, both the alt-right and far-left groups incited violence in the weeks leading up to the Charlottesville rally and called for action among their supporters. A tweet from one of the alt-right groups read 'The left is preparing lynch mobs to descend on the Unite The Right rally in Charlottesville, VA... This is going to be fun.' [22].

Beyond the Charlottesville rally, the effect of Jihadist terrorism and Islamophobic attacks on hate speech has also been measured on Twitter and Reddit [23]. One study measured hate speech over a period of 19 months shortly after 13 extremist attacks. It was found that, following Islamist terrorist attacks, hate speech targeting Muslims, particularly those advocating violence, increased more after terror attacks compared to a counterfactual simulation [23]. At the same time, an increase in messages countering hate speech (e.g. defending Muslims) after Islamist terrorist attacks was observed [23]. In contrast, following Islamophobic attacks, an increase in hate speech targeting Muslims was not found, with the exception of messages posted after the 2017 Finsbury Park Mosque attack [23].

Other investigations have examined changes in language over time in the context of political change. For example, hate speech and white nationalist rhetoric during the 2016 US elections were studied on Twitter [24]. The aim was to empirically examine a potential rise in hate speech as a result of the Donald Trump's divisive election campaign [24]. That study compared tweets referring to Donald Trump and Hillary Clinton between June 15, 2015 (the day after the Trump candidacy was announced) and June 15, 2017, and compared these with a random sample of tweets from American Twitter users. Hate speech and white nationalist language was measured with a dictionary approach using Hatebase, the Racial Slur Database, and the Anti-Defamation League's database of white-nationalist language[1] [24]. The tweets were examined by means of an interrupted time series analysis, showing a spike in hate speech in the

---
[1] See https://hatebase.org/, http://www.rsdb.org/ and https://www.adl.org/education-and-resources/resource-knowledge-base/hate-symbols

Trump dataset following the imposed travel ban in early 2017. No further significant lasting increases of hate speech or white nationalist language were observed in the Trump, Clinton, or random sample data [24].

III. AIMS OF THIS PAPER

The present study takes a closer look at alternative content creators 'who range in ideology from mainstream libertarian to openly white nationalist' [7] (hereafter, 'alternative group'), by examining YouTube video transcripts extracted from channels by these individuals. These video transcripts are compared to those from YouTube channels whose political orientation can be considered more progressive (hereafter, 'progressive group').

This paper has two aims. First, we compare language use *between* the alternative and progressive group in a sixteen-week timeframe surrounding the Charlottesville rally. Second, we assess whether the rally had an effect on language use *within* the two groups. For the alternative group, we do not postulate any directional hypotheses about changes in language use. Nevertheless, in light of the social movement and social identity performance literature, we expect to see changes in social identity performance after the rally reflected in language use on YouTube. For the progressive group, we do not claim that the channels studied act as a social movement, and thus we have no expectations of social identity performance. However, we are interested in seeing whether the channels lend any discursive opportunities to the alt-right through language use in their videos, thereby potentially fuelling increased social identity performance on part of the alternative group.

The first aim is addressed through structural topic modelling, in order to present the prevalence of topics between the two groups across time. The second aim is addressed using a word frequency approach, in which we examine the frequency of common phrases before and after the Charlottesville rally, searching for sudden increases or decreases as a result of the rally.

IV. METHOD

*A. Data availability statement*

Supplemental materials, data and code to reproduce the analysis are available on the Open Science Framework: https://osf.io/yedt7/?view_only=ec235ff877d24731ad5140e24c7a6cdd

*B. Data*

*1) Channel selection*

YouTube channels were selected for analysis from two main sources. First, we drew from the list of 65 YouTube users referred to as the 'Alternative Influence Network' in the 2018 Data & Society report[2] on political influencers [8]. Based on this list, we searched for a designated YouTube channel for each individual. If an individual did not have a designated YouTube channel or their channel was no longer available, we searched for the individual's name through the YouTube search function. For example, videos featuring Alex Jones (who no longer has a designated channel) were obtained through the search query 'alex jones full show'. The group of alternative YouTube channels consisted of 56 channels and search queries to be considered for transcript retrieval. Second, for the comparison group of progressives, we drew from two online lists of progressive YouTube channels[3]. Since the lists referred to specific existing channels, search queries for specific persons were not necessary. In total, 13 progressive channels were considered for transcript retrieval. For all channels and search queries, we retrieved the URLs for all available videos.

*2) Transcript retrieval*

The method for retrieving YouTube video transcripts follows the procedure of related research [25, 26]. In order to retrieve the transcripts, a Python script was written using www.downsub.com to obtain XML-encoded transcripts. The transcripts were either automatically generated by YouTube or manually added by the YouTube user. In some cases, no transcript was available, because users disabled the transcript availability. XML-tags and time-stamps were removed, resulting in a single, non-punctuated string for each video transcript.

---

[2] https://datasociety.net/wp-content/uploads/2018/09/DS_Alternative_Influence.pdf
[3] http://the2020progressive.com/top-13-progressive-news-shows-youtube/ and https://medium.com/@tejazz89/top-5-youtube-channels-to-follow-if-you-are-a-true-progressive-ee2abc78d58f

*3) Data cleaning*

Videos that contained fewer than 100 words were not considered for analysis. Using R software, each video was checked for English language, and was excluded if it contained fewer than 50% English words. Videos were also excluded if they contained fewer than 90% ASCII characters. The video transcript strings were lower-cased and stopwords, unnecessary whitespace or punctuation were removed using the R packages *tidytext* [27], *tm* [28] and *qdap* [29].

*4) Sample*

For the purpose of analysing potential effects of the rally on language use, we selected videos from the dataset that were posted between eight weeks prior to the Charlottesville rally, to eight weeks after the rally. Descriptive statistics for this sample are given in Table I.

**Table I** Descriptive statistics video sample

|  | **Alternative** | **Progressive** |
|---|---|---|
| Total videos | 2,684 | 4,458 |
| Total word count | 3,868,744 | 2,804,703 |
| Word count | Mean: 1,448 (SD = 1,612) | Mean: 632 (SD = 860) |
|  | Min: 34, Max: 12,085 | Min: 33, Max: 11,645 |
| View count | Mean: 120,639 (SD = 234,513) | Mean: 24,787 (SD = 47,468) |
|  | Min: 2, Max: 17,340,303 | Min: 12, Max: 2,475,766 |

*C. Analysis plan*

*1) Structural topic model*

To assess the differences in language use between the alternative and progressive groups, we construct structural topic models. This method can be used to automatically extract underlying latent topics in a corpus [30, 31]. A common approach is Latent Dirichlet Allocation [30], a probabilistic model which is based on the assumption that a piece of text consists of a mix of topics, which in turn are a mix of words with probabilities of belonging to a topic [30, 31]. A structural topic model is a type of probabilistic model, with the added benefit that one can incorporate document-level covariates (e.g., document author, political orientation, date) and assess whether these covary with topic prevalence (i.e. how much of a document is associated with a topic) [31].

We first construct structural topic models using unigrams (e.g., 'trump') as unit of analysis, followed by models with bigrams (e.g., 'donald trump') as unit of analysis. We include group (alternative vs. progressive) and date (from 15 June to 7 October 2017, 16 weeks surrounding the rally) as covariates for topic prevalence. Unigram and bigram topic models are fit with a varying number of topics[4], after which we select the best fitting model based on the trade-off between semantic coherence and exclusivity [31, 32], two metrics frequently used to assess whether a topic is semantically useful [31, 33]. Semantic coherence is a measure of the co-occurrence of highly probable words in a topic, and has been shown to correlate with expert judgments of topic quality [32]. It has been proposed that a measure of exclusivity of words to topics is needed to further determine topic quality, otherwise several topics may be represented by the same highly probable words, if one relies on semantic coherence alone. Exclusive topics are made up of words that have a high probability under one topic, but a low probability under other topics [33].

After selecting a model, we present topics for which at least one of the covariates showed a significant effect, in order of total expected topic proportion for the corpus. We indicate whether group, date, or an interaction between the two significantly covaried with topic prevalence. Based on manual inspection of frequent and exclusive topic words [31, 34], we assign labels to topics where possible.

*2) Word frequency*

To examine possible changes in word frequency within the alternative and progressive group as a result of the rally, we compute the frequency of all bigrams for each week in both groups separately. By dividing these values by the total number of bigrams for each day, we obtain the daily proportion for each bigram. Thereafter, we can assess whether there is a structural breakpoint in the proportion of each bigram as a result of the rally. This is done by means of the Chow test [35, 36], with which we determine whether a breakpoint in the intercept and slope occurred at the time of the rally. In order to do so, we test for the equality between a model of bigram proportions before the Charlottesville rally, and a model of bigram proportions after the rally. In both models, the proportion of each bigram is represented as a function of Date (day on which the proportion was measured, between 15 June and 7 Oct 2017). We compute an F-value for the equality between the two models for each bigram, and report those which are found to differ significantly pre- and post-rally. In addition, we present associated intercept and slope changes.

---
[4] 20, 40, 50, 60, 70, 80, and 90 topics

## V. RESULTS

### A. *Structural topic model*

For unigrams, we decided on a structural topic model with 40 topics based on examination of semantic coherence and exclusivity for each model (see supplemental materials for results with different numbers of topics). Thereafter, we found that the covariates (group and date) were significant for the prevalence of thirteen topics, both separately and as an interaction ($p < 0.01$). Figure I shows the topics for which group, date, or both significantly covaried with topic prevalence. We assigned labels (e.g. 'Obamacare') based on examination of frequent and exclusive words.

In Figure II we present three topics (feminism, swearing, and climate) in more detail, in terms of their prevalence over time. These topics were selected because they differed most between groups (i.e. the three largest regression coefficients for covariate 'group'). The topic relating to feminism was discussed to a larger extent by alternative channels, albeit decreasing over time. A similar pattern could be observed for the topic related to swearing, with the alternative group using swearwords more, but with a decreasing trend. In contrast, both groups discussed the climate and weather with increasing prevalence, although this trend was stronger in the progressive group.

**Figure I** Unigram topic proportions

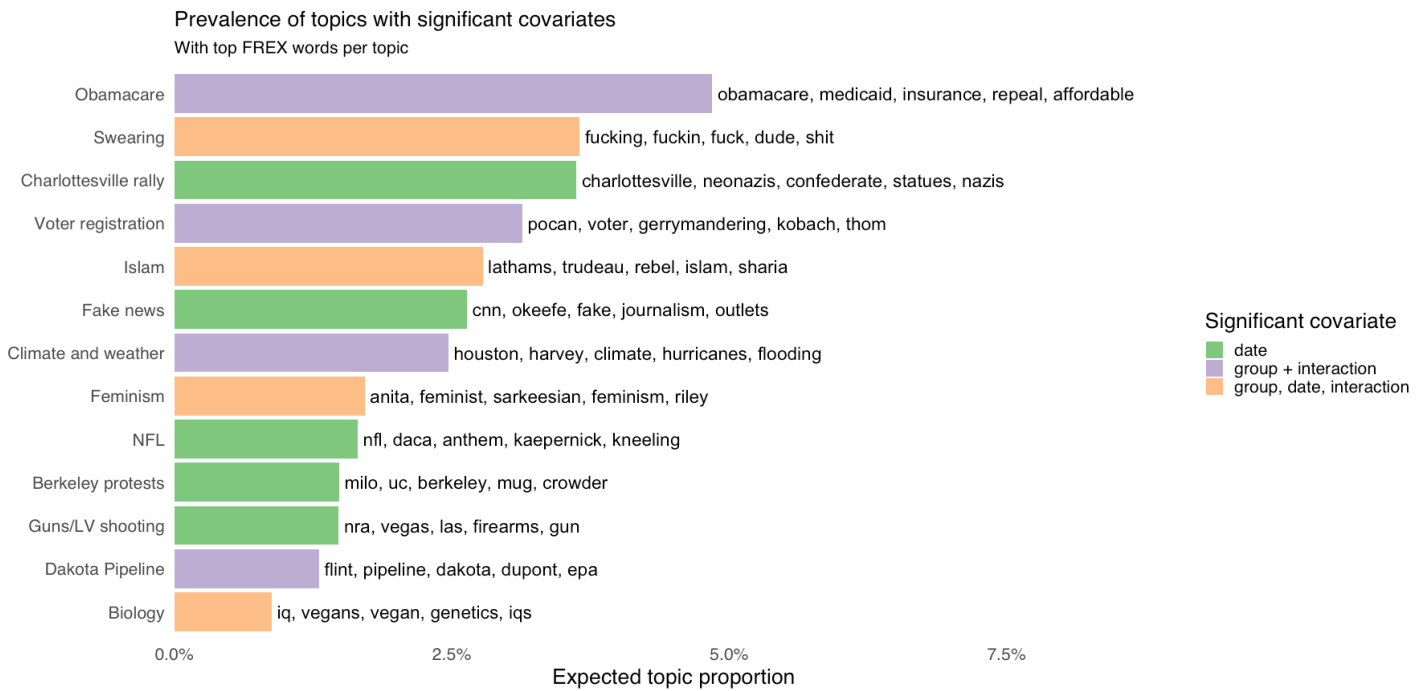

**Figure II** Selected unigram topics over time

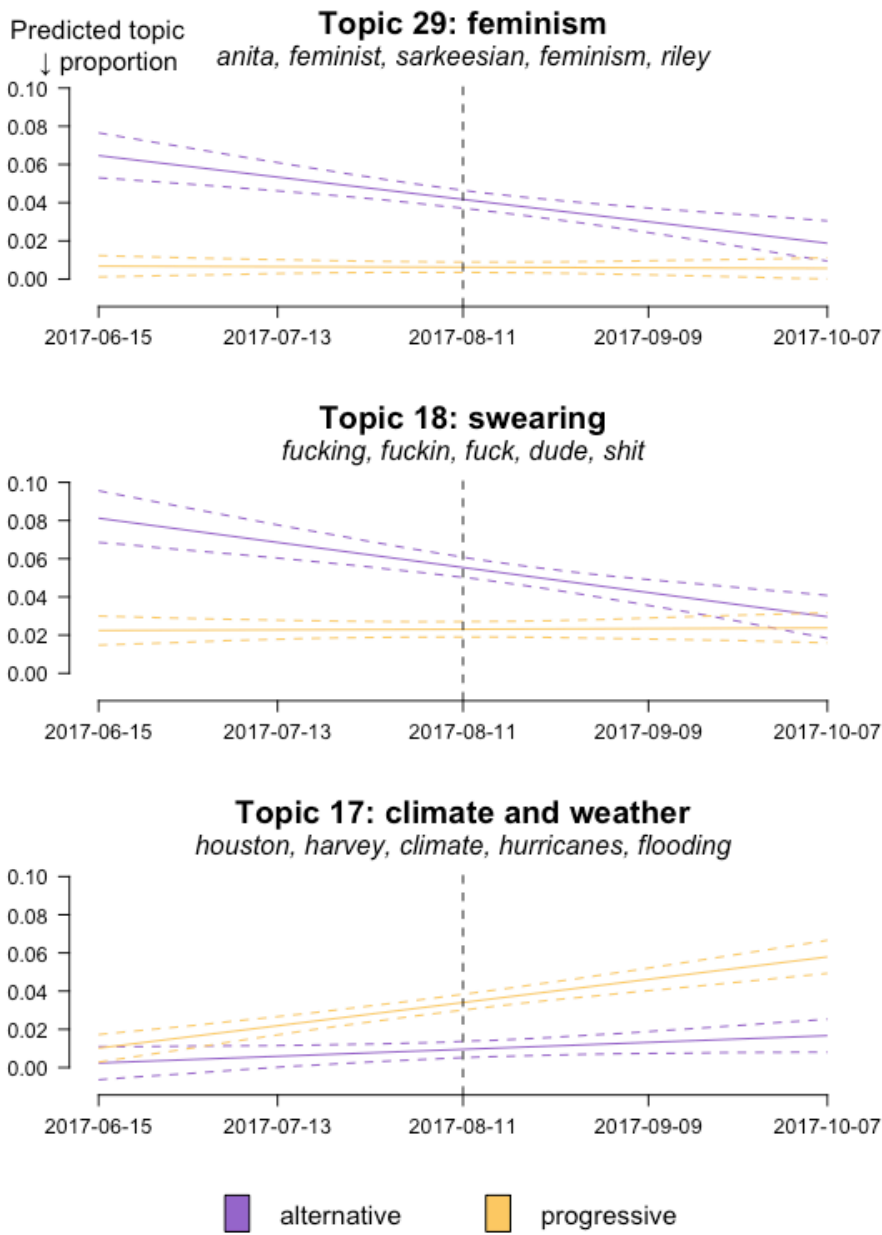

*Note.* The y-axis represents the proportion of documents related to the topic in question at each time point. The vertical dotted line in the middle of the plot represents (the first day of) the rally.

For bigrams, we decided on the model with 50 topics based on examination of semantic coherence and exclusivity for each model (see OSF for results with different numbers of topics). Figure III shows the topics for which at least one of the covariates was significant ($p < 0.01$). Again, we assigned topic labels based on frequent and exclusive words. However, for some instances no interpretable topic emerged, and these are indicated by topic number (e.g. 'Topic 31').

    Figure IV shows the three topics for which the covariate group had the largest coefficient, namely race, free speech, and healthcare financing. The bigram topic relating to race was discussed by alternative channels with increasing frequency over the time period surrounding Charlottesville, with little to no such discussion in the progressive group. Similarly, the prevalence of the topic 'free speech' increased over the same time period in the alternative group, but no pronounced change was observed in the progressive group. The progressive group discussed healthcare financing to a larger extent than the alternative channels, although the trend decreased over time until there was no longer a large difference between the groups towards the end of the timeframe.

**Figure III** Bigram topic proportions

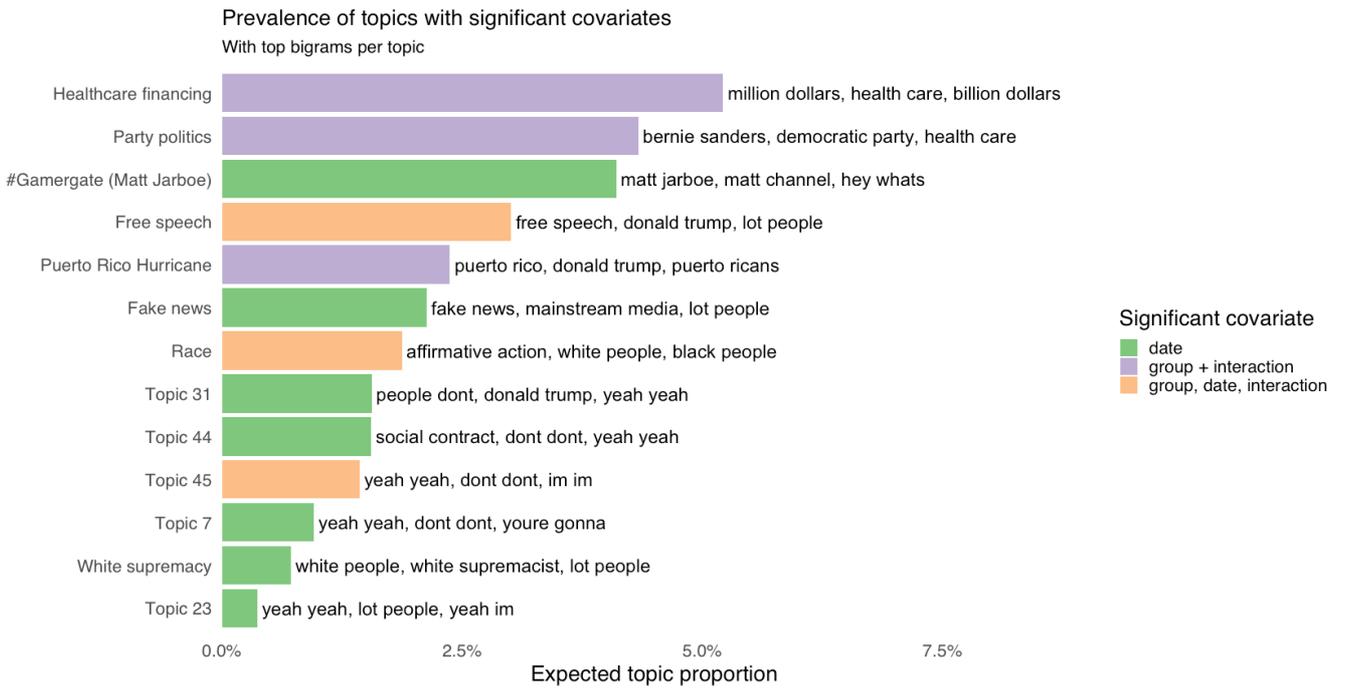

**Figure IV** Selected bigram topics over time

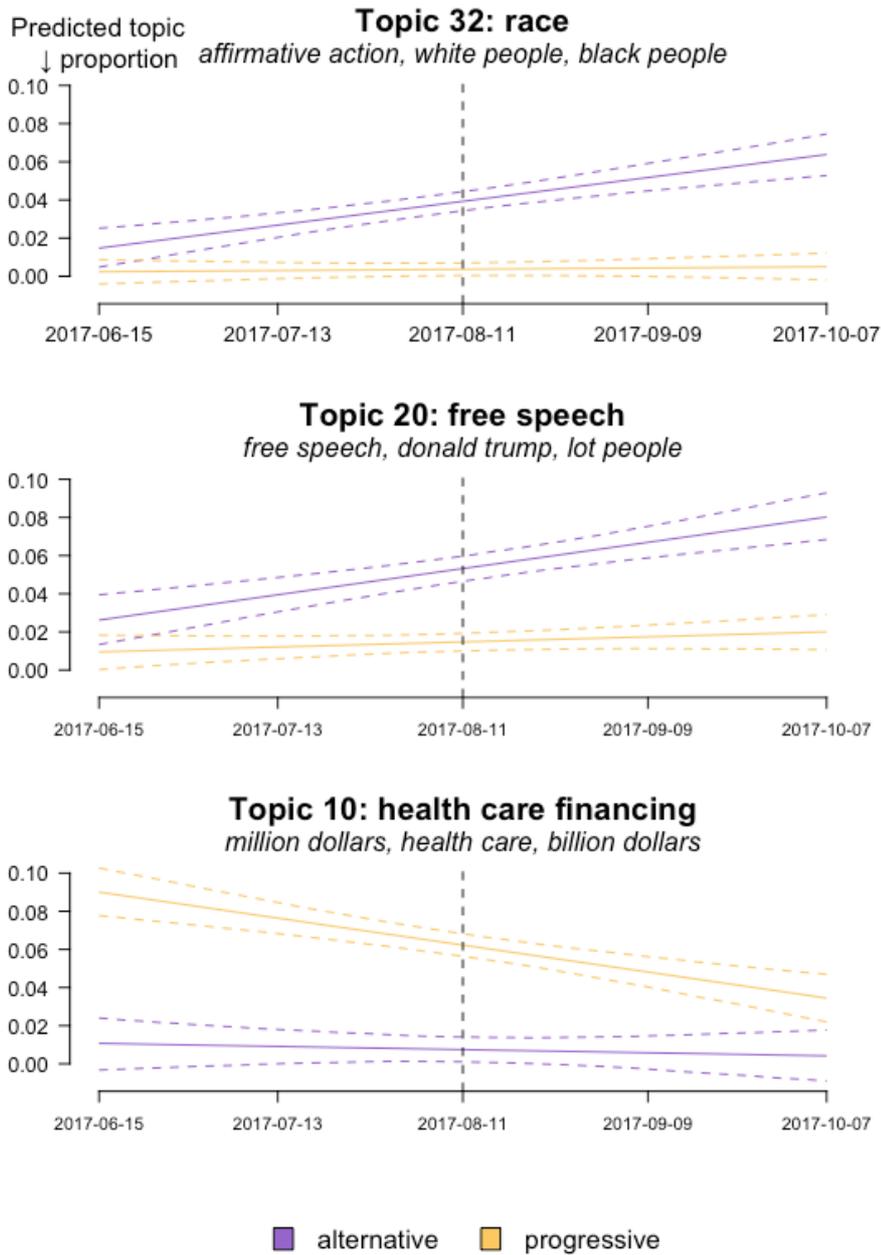

B. *Word frequency approach*

We show the ten bigrams for which the Chow test *F* statistic, indicative of a joint breakpoint in intercept and slope, was largest[5], in the alternative group (Table II) and progressive group (Table III). We also show the direction and magnitude of intercept and slope changes after the rally; please note that slope changes were very minimal (albeit significant) and therefore have been multiplied by 10,000 for interpretability. Among the top ten bigrams with breakpoints for both groups, the majority relate to the rally itself, such as 'white nationalist' and 'happen charlottesville'. Note that some bigrams showed a breakpoint in both groups, namely, 'white nationalist', 'happen charlottesville', 'charlottesville virginia', and 'neonazi white'. In the alternative group, several bigrams unrelated to the rally (e.g. 'hit bell', 'video bitcoin') also exhibit strong breakpoints. In the progressive group, only one bigram with a strong breakpoint in the top ten seems to be unrelated to the rally, namely 'hurricane maria'. In order to further illustrate the bigram proportion breakpoints, we show the progression of the first three (based on the magnitude of the Chow test *F*) bigrams for the alternative group (Figure V) and the progressive group (Figure VI).

---

[5] Further bigrams that exhibited breakpoints are available in the supplemental materials.

In both groups, the proportion of the bigrams depicted significantly increases in terms of intercept, with slight (negative) changes in slopes.

**Table II** Ten bigrams with largest Chow test (*F*) statistic in alternative group

| Bigram | Chow test (*F*)[a] | Effect size *d* | Intercept change | Slope change[b] |
|---|---|---|---|---|
| white nationalist | 42.84 | 0.89 | 1.12 | -0.64 |
| happen charlottesville | 38.09 | 0.84 | 0.20 | -0.11 |
| hit bell | 35.47 | 0.81 | 0.26 | -0.15 |
| video bitcoin | 31.85 | 0.76 | 0.20 | -0.11 |
| subscribe hit | 28.92 | 0.73 | 0.24 | -0.14 |
| charlottesville virginia | 28.10 | 0.72 | 0.11 | -0.06 |
| nazi flag | 26.86 | 0.70 | 0.19 | -0.11 |
| descript patreon | 26.70 | 0.70 | 0.19 | -0.11 |
| neonazi white | 26.00 | 0.69 | 0.11 | -0.06 |
| begin video | 25.90 | 0.69 | 0.33 | -0.19 |

*Notes.* [a]For all coefficients (F, intercept and slope changes): $p < 0.001$
[b]Slope change estimates have been multiplied by 10,000 for interpretability

**Table III** Ten bigrams with largest Chow test (*F*) statistic for progressive group

| Bigram | Chow test (*F*)[a] | Effect size *d* | Intercept change | Slope change[b] |
|---|---|---|---|---|
| white nationalist | 42.84 | 0.89 | 1.12 | -0.64 |
| happen charlottesville | 38.09 | 0.84 | 0.20 | -0.11 |
| hit bell | 35.47 | 0.81 | 0.26 | -0.15 |
| video bitcoin | 31.85 | 0.76 | 0.20 | -0.11 |
| subscribe hit | 28.92 | 0.73 | 0.24 | -0.14 |
| charlottesville virginia | 28.10 | 0.72 | 0.11 | -0.06 |
| nazi flag | 26.86 | 0.70 | 0.19 | -0.11 |
| descript patreon | 26.70 | 0.70 | 0.19 | -0.11 |
| neonazi white | 26.00 | 0.69 | 0.11 | -0.06 |
| begin video | 25.90 | 0.69 | 0.33 | -0.19 |

*Notes.*[a]For all coefficients (*F*, intercept and slope changes)*: p < 0.001*
[b]Slope change estimates have been multiplied by 10,000 for interpretability

**Figure V** Observed proportion of three bigrams with largest F-value in alternative group

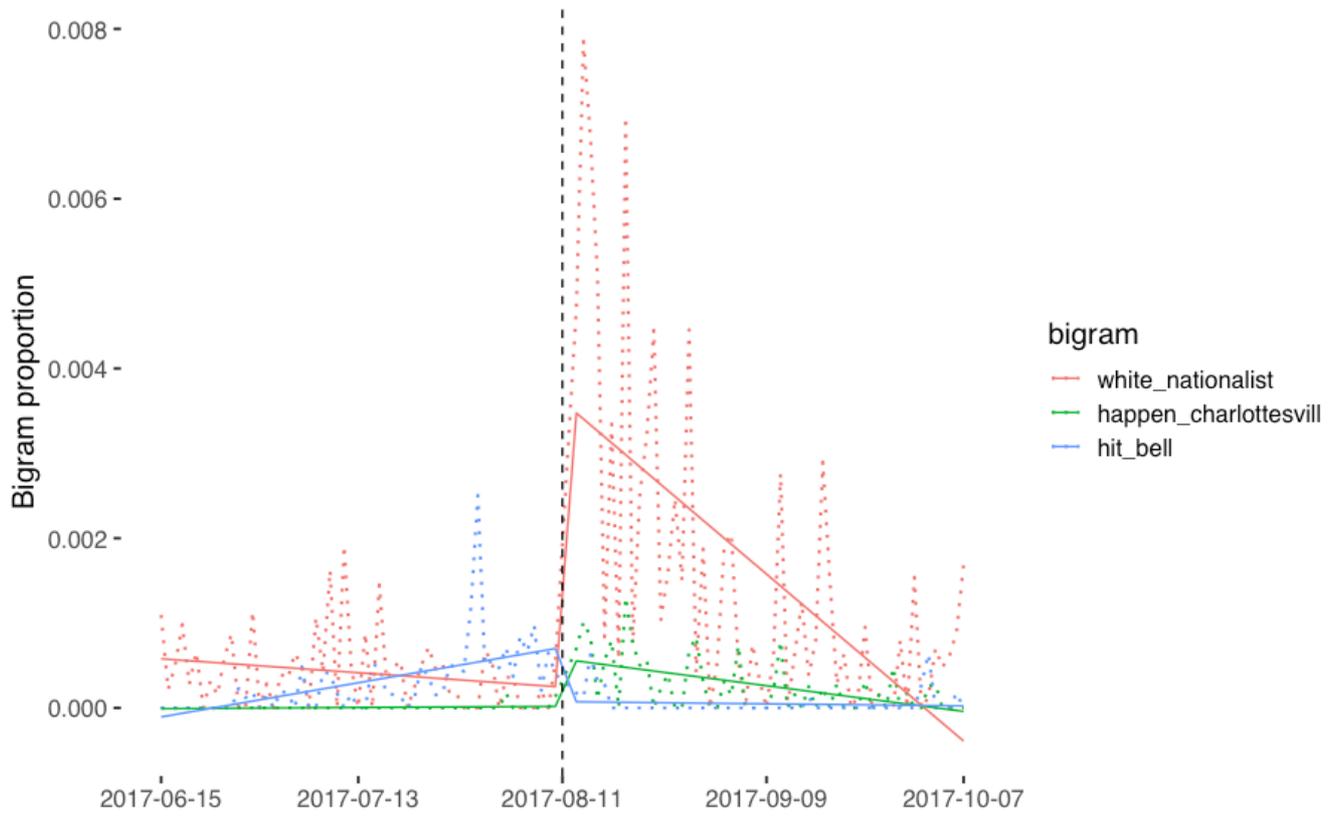

**Figure VI** Observed proportions of three bigrams with largest with largest F-value in the progressive group

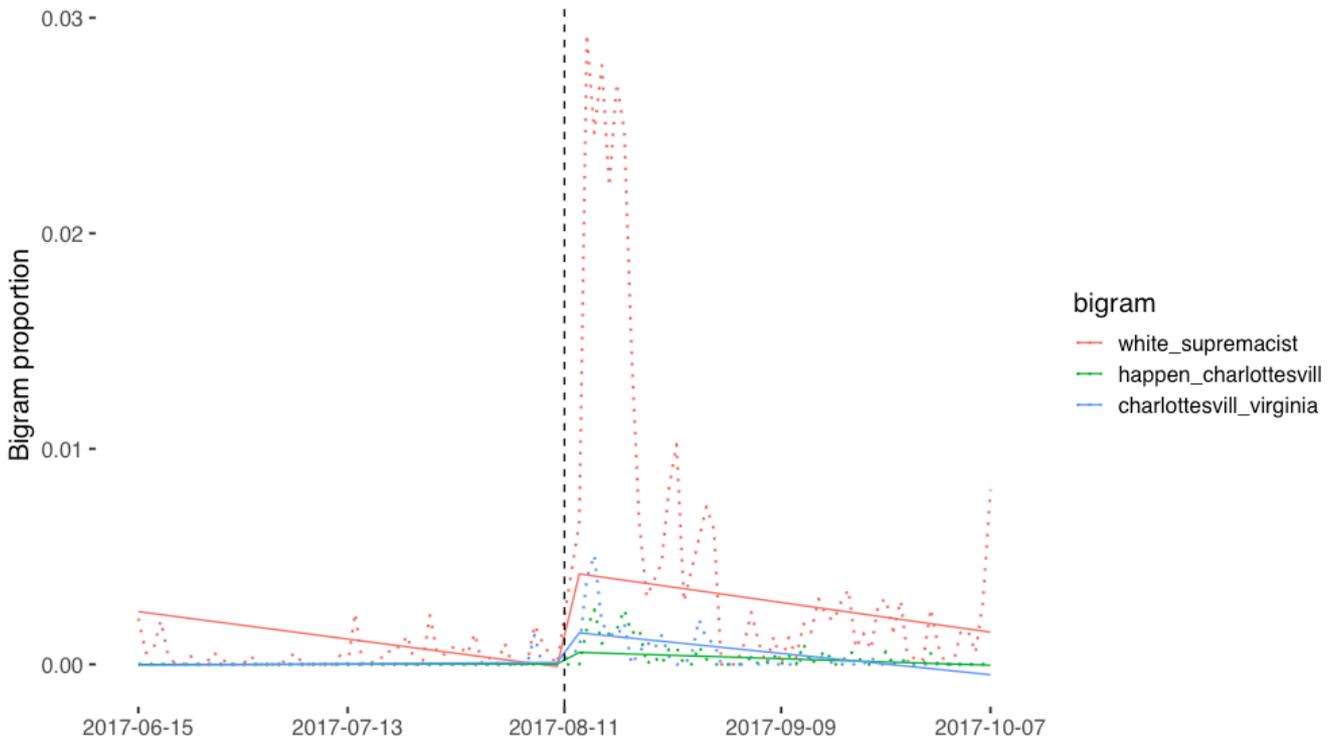

## VI. DISCUSSION

The current study examined language use for alternative and progressive YouTube channels around the time of the Charlottesville 'Unite the Right' rally. The aims of this paper were to compare language use between the groups surrounding the rally, and to assess whether the rally had an effect on language use within the two groups. We examined language use in both groups in terms of structural topic models, and searched for structural breakpoints in a change of content as a result of the rally. We consider the outcome of both approaches in turn, with interpretations of the results in light of social movement theory.

### A. Differences between alternative and progressive channels

The first line of inquiry examined whether there were structural differences in the prevalence of topics between groups. This analysis illustrates the matters discussed in videos throughout this period in the two groups. Perhaps unsurprisingly, both groups discussed several topics related to politics and current events (e.g. healthcare and climate change). The prevalence of discussions on specific events that occurred between mid-June and mid-October 2017 indeed varied across time, such as the Charlottesville rally, the national anthem protests in the NFL [37], the Las Vegas shooting [38], and pro- and anti-Trump protests in Berkeley [39].

We also observed that the prevalence of some topics covaried with the political orientation of channels (alternative or progressive). For instance, topics that may be loosely associated with the 'ideology' of the alt-right were found to differ between groups in our corpus, such as (an aversion to) feminism, designating media outlets as 'fake news', white nationalism, and leveraging free speech rights [11, 12]. Indeed, the concept of free speech has frequently been linked to the alt-right and white nationalism, where the right to free speech is used to "advance racist and sexist ideas" [40]. In a similar vein, discussions relating to (anti-)feminism have also been linked to the far right [41], a further potential indicator of expressing social norms within this group. Interestingly, we also observed a difference in language use in terms of swearing, the prevalence of which was significantly higher for alternative influencers (based on the coefficient for the covariate group). Swearing may be a way of conforming to a social group, and our results suggest that this kind of language is more common among alternative than progressive YouTube channels. In short, the structural topic models indeed show that there are differences in topics between alternative and progressive YouTube channels, and that these differences also vary over time. Some of these patterns in topics may support previous claims that the alt-right behaves as a social movement [9].

*B. Effects of the rally within alternative and progressive channels*

The word frequency approach showed the rally had an effect on language use within the two groups, illustrated by several breakpoints in bigram proportions that coincided with the Charlottesville rally. Unsurprisingly, the use of words relating to the rally (e.g. confederate monument, white nationalist, white supremacist) increased at this point. While the proportions of these bigrams all exhibited sudden increases, the mentions did decrease over time in the post-rally timeframe. This possibly reflects a 'natural' descending trend for discussions of an event as time progresses, which potentially adds to the justification of measuring bigram proportions over time to assess reactions to events in language.

Although there was some overlap between groups in bigram use, it also appears that both groups discussed the events in a different light. The progressive group increasingly mentions 'white supremacists' after the rally, whereas the alternative group increasingly mentions 'white nationalists'[6]. These differences in terminology seem to reflect a more general divide between groups. Indeed, 'white supremacists' is a term preferred by people who study or condemn the movement, but the term is not preferred among the extreme right itself [12]. Among the alt-right, the preferred term is 'white nationalist', which indeed emerges from our data [12]. This preference relates to the wish to establish separate white *nations*, in contrast to multiracial nations where whites are the dominant ('supreme') group [12]. One could argue that this difference in terminology may reflect increased expressions of in-group (alt-right) norms, an aspect of social identity performance.

Further breakpoints observed in the progressive group refer to several details related to the rally, such as the confederate statue of Robert Lee, the removal of which gave rise to the Charlottesville rally [42]. A strong increase within progressive post-rally videos was observed for the mention of counter-protestors, highlighting potential condemnation of the rally and the violence that ensued against counter-protestors [43]. Interestingly, none of these details appear in the top ten of breakpoints for the alternative group. We do not propose that these patterns in language use provide evidence for social identity performance on part of the progressive group, as we studied a user-generated and highly heterogenous list of channels, for which, in contrast to the alternative group, no claims have been made that they form a social movement. However, mentions of the rally on part of the progressive group may have lent further discursive opportunities to the alternative group, which may in turn have fuelled social identity performance on part of the latter group [19, 20]. That is, by mentioning and even condemning the alt-right rally, the progressive group lends further resonance and visibility to the movement [20].

Interestingly, a large number of the top ten bigrams in the alternative group for which a breakpoint was observed did not relate to the rally, but to the promotion of YouTube channels, for instance urging viewers to subscribe to a channel, enable notifications, or donate to Patreon, a platform where content creators can crowdsource donations [44]. While further examination of the contexts in which these calls are made will be needed, the fact that (positive) breakpoints (in intercept) appear at the time of the rally may be a sign of mobilising others, for example to show their support for the alternative channels and related movements. Indeed, if these calls are a direct result of the rally, they may serve as evidence of increased social identity performance, attempting to strengthen a movement. In short, the examination of bigram proportion breakpoints showed that the Charlottesville rally did seem to have an effect on language use in both groups separately. However, further research will be needed to examine the exact contexts that gave rise to the changes and to confirm whether they serve as direct evidence of social identity performance.

*C. Limitations*

The current study is not without limitations. First, the data sources that we have drawn on for the YouTube videos were unbalanced in nature. Although the progressive sample consisted of more videos than the alternative sample, the progressive sample was less diverse than the alternative sample. This was due to the fact that we were able to attain a large list of alternative YouTube channels through the Data & Society report, but to the best of our knowledge, no similar report or collection exists for progressive YouTube channels. Furthermore, the two groups also differed in terms of view counts and video length, both factors which may have impacted on language use. In addition, while the alternative channels were drawn from a research report, the list of progressive channels were drawn from user-generated online lists. Future research may be aimed at curating an expert-verified or crowd-sourced dataset of channels with different political biases[7].

Topic modelling involves several decisions on part of the researcher. For instance, various approaches exist for selecting the number of topics for a model, with no consensus in the research community [31]. Furthermore, assigning labels to topics is based on the interpretation of the researcher, with decisions highly sensitive to human bias. Nevertheless, we provide alternative models (with different numbers of topics) and further terms associated with topics in the supplemental materials, for the reader to examine the outcome of our analyses, giving way to alternative

---

[7] Similar to https://mediabiasfactcheck.com/ and https://www.allsides.com/media-bias/media-bias-ratings but for YouTube channels

explanations. Furthermore, some topics were difficult to interpret, mostly because they were composed of parts-of-speech with little meaning, or because the words did not form a coherent topic, and merely consisted of words that were used in the same way. In those cases, we refrained from assigning labels.

The word frequency approach measured the prevalence of bigrams, which are restricted in the sense that additionally meaningful n-grams may not have been considered. That is, polarity words or adjectives that preceded bigrams may not have been captured. This issue may be solved in future by using trigrams, although relevant n-grams that occur even further away from the keyword will still not be captured. As has been raised in the discussion, the breakpoints we observed only show that there was a change in frequency (proportion) of a bigram, and say nothing about the context in which bigrams occurred. For example, mentions of 'white nationalist' may have appeared in a negative context in the progressive group, and a positive context in the alternative group, but further analyses will be needed to make such claims.

*D. Outlook*

It can be argued that understanding of changes in language use of potentially violent groups on social media may be of particular interest to policy makers and security officials aiming to prevent or de-escalate violence. Future research may focus on extending the present approach to measuring changes in language over time on other social media platforms where alt-right supporters are active, such as 8Kun and Gab. It may also be of interest to measure concepts other than topics and n-gram frequencies, such as hate speech and abusive language, in response to the Charlottesville rally and perhaps other events of interest. Although it is beyond the scope of the current paper, a follow-up study of the specific contexts in which certain topics and n-grams occur may be interesting. For example, is the sentiment regarding 'white people' or 'feminism' negative or positive in polarity?

## VII. CONCLUSION

Following the violent rally in Charlottesville, the alt-right received significant attention in the media and public discourse. As a result, we expected to see differences in social identity performance on part of the alt-right movement, which was measured through examining language use. Contrasting a unique dataset of YouTube video transcripts from alternative, right-leaning channels to progressive, left-leaning channels, the present investigation indeed observed differences in language within and between the alternative and progressive groups. Results potentially reflect changes in social identity performance after the rally, as well as differences between the two groups more generally.

## VIII. DECLARATIONS

On behalf of all authors, the corresponding author states that there is no conflict of interest.

This project has received funding from the European Research Council (ERC) under the European Union's Horizon 2020 research and innovation programme (Grant Agreement No. 758834).

REFERENCES

1. Hughes, T. (2018). A year after Charlottesville rally, white nationalists enter mainstream conversation. *USA Today*. Retrieved June 14, 2019, from https://eu.usatoday.com/story/news/2018/08/09/white-nationalists-riding-recognition-push-views-into-mainstream/935159002/

2. Yan, H., & Sayers. (2017). Virginia governor on white nationalists: They should leave America. *CNN*. Retrieved July 3, 2019, from https://www.cnn.com/2017/08/13/us/charlottesville-white-nationalist-rally-car-crash/index.html

3. Anti-Defamation League. (2017). Murder and Extremism in the United States in 2017. *Anti-Defamation League*. Retrieved July 3, 2019, from https://www.adl.org/resources/reports/murder-and-extremism-in-the-united-states-in-2017

4. Anti-Defamation League. (2018). Murder and Extremism in the United States in 2018. *Anti-Defamation League*. Retrieved July 3, 2019, from https://www.adl.org/murder-and-extremism-2018

5. Hine, G., Onaolapo, J., Cristofaro, E. D., Kourtellis, N., Leontiadis, I., Samaras, R., … Blackburn, J. (2017). Kek, Cucks, and God Emperor Trump: A Measurement Study of 4chan's Politically Incorrect Forum and Its Effects on the Web. In *Proceedings of the Eleventh International AAAI Conference on Web and Social Media* (p. 10).

6. Zannettou, S., Bradlyn, B., De Cristofaro, E., Kwak, H., Sirivianos, M., Stringini, G., & Blackburn, J. (2018). What is Gab: A Bastion of Free Speech or an Alt-Right Echo Chamber. In *Companion Proceedings of the The Web Conference 2018* (pp. 1007–1014). Republic and Canton of Geneva, Switzerland: International World Wide Web Conferences Steering Committee. https://doi.org/10.1145/3184558.3191531

7. Ellis, E. G. (2018, September 19). The Alt-Right Are Savvy Internet Users. Stop Letting Them Surprise You. *Wired*. Retrieved from https://www.wired.com/story/alt-right-youtube-savvy-data-and-society/

8. Lewis, R. (2018). Broadcasting the Reactionary Right on YouTube. *Data & Society*. Retrieved from: https://datasociety.net/wp-content/uploads/2018/09/DS_Alternative_Influence.pdf

9. Hodge, E., & Hallgrimsdottir, H. (2019). Networks of Hate: The Alt-right, "Troll Culture", and the Cultural Geography of Social Movement Spaces Online. *Journal of Borderlands Studies*, 1–18. https://doi.org/10.1080/08865655.2019.1571935

10. Hawley, G. (2018). *The Alt-Right: What Everyone Needs to Know®*. Oxford University Press.

11. Nagle, A. (2017). *Kill all normies: the online culture wars from Tumblr and 4chan to the alt-right and Trump*. Winchester, UK ; Washington, USA: Zero Books.

12. Hawley, G. (2017). *Making Sense of the Alt-Right*. Columbia University Press.

13. Tilly, C. (1993). Contentious Repertoires in Great Britain, 1758–1834. *Social Science History*, *17*(2), 253–280. https://doi.org/10.1017/S0145553200016849

14. Serwer, A. (2018, August 10). The White Nationalists Are Winning. *The Atlantic*. Retrieved from https://www.theatlantic.com/ideas/archive/2018/08/the-battle-that-erupted-in-charlottesville-is-far-from-over/567167/

15. Shear, M. D., & Haberman, M. (2018, January 20). Trump Defends Initial Remarks on Charlottesville; Again Blames 'Both Sides.' *The New York Times*. Retrieved from: https://www.nytimes.com/2017/08/15/us/politics/trump-press-conference-charlottesville.html

16. Atkinson, D. C. (2018). Charlottesville and the alt-right: a turning point? *Politics, Groups, and Identities*, *6*(2), 309–315. https://doi.org/10.1080/21565503.2018.1454330

17. Klein, O., Spears, R., & Reicher, S. (2007). Social Identity Performance: Extending the Strategic Side of SIDE. *Personality and Social Psychology Review*, *11*(1), 28–45. https://doi.org/10.1177/1088868306294588

18. Simon, B., Trötschel, R., & Dähne, D. (2008). Identity affirmation and social movement support. *European Journal of Social Psychology*, *38*(6), 935–946. https://doi.org/10.1002/ejsp.473

19. Koopmans, R., & Muis, J. (2009). The rise of right-wing populist Pim Fortuyn in the Netherlands: A discursive opportunity approach. *European Journal of Political Research*, *48*(5), 642–664. https://doi.org/10.1111/j.1475-6765.2009.00846.x

20. Koopmans, R., & Olzak, S. (2004). Discursive Opportunities and the Evolution of Right-Wing Violence in Germany. *American Journal of Sociology*, *110*(1), 198–230. https://doi.org/10.1086/386271

21. Peltz, J. (2017, August 14). Protests, vigils around US decry white supremacist rally. *NY Daily News*. Retrieved July 3, 2019, from https://web.archive.org/web/20170814021328/http://www.nydailynews.com/newswires/new-york/protests-vigils-decry-white-supremacist-rally-article-1.3408420

22. Klein, A. (2019). From Twitter to Charlottesville: Analyzing the Fighting Words Between the Alt-Right and Antifa. *International Journal of Communication*, *13*(0), 22.

23. Olteanu, A., Castillo, C., Boy, J., & Varshney, K. R. (2018). The Effect of Extremist Violence on Hateful Speech Online. In *Twelfth International AAAI Conference on Web and Social Media*. Presented at the Twelfth International AAAI Conference on Web and Social Media. Retrieved from https://www.aaai.org/ocs/index.php/ICWSM/ICWSM18/paper/view/17908

24. Siegel, A. A., Nikitin, E., Barbera, P., Sterling, J., Pullen, B., Bonneau, R., … Tucker, J. A. (2018). Measuring the prevalence of online hate speech, with an application to the 2016 U.S. election. Retrieved from: https://smappnyu.org/wp-content/uploads/2018/11/Hate_Speech_2016_US_Election_Text.pdf

25. Kleinberg, B., Mozes, M., & van der Vegt, I. (2018). Identifying the sentiment styles of YouTube's vloggers. In *Proceedings of the 2018 Conference on Empirical Methods for Natural Language Processing*. Retrieved from http://arxiv.org/abs/1808.09722

26. Soldner, F., Ho, J. C., Makhortykh, M., van der Vegt, I. W. J., Mozes, M., & Kleinberg, B. (2019). Uphill from here: Sentiment patterns in videos from left- and right-wing YouTube news channels. In *Proceedings of the Third Workshop on Natural Language Processing and Computational Social Science* (pp. 84–93). Minneapolis, Minnesota: Association for Computational Linguistics. Retrieved from: https://www.aclweb.org/anthology/W19-2110

27. Silge, J., & Robinson, D. (2019). *Text Mining with R*. Retrieved from https://www.tidytextmining.com/

28. Feinerer, I., & Hornik, K. (2018). *tm: Text Mining Package*. Retrieved from https://CRAN.R-project.org/package=tm

29. Rinker, T. W. (2019). *Quantitative Discourse Analysis Package*. Retrieved from https://cran.r-project.org/web/packages/qdap/citation.html

30. Blei, D. M. (2003). Latent Dirichlet Allocation. *Journal of Machine Learning Research*, 30.

31. Roberts, M. E., Stewart, B. M., & Tingley, D. (2014). stm: R Package for Structural Topic Models. *Journal of Statistical Software*, 41.

32. Mimno, D., Wallach, H., Talley, E., Leenders, M., & McCallum, A. (2011). Optimizing Semantic Coherence in Topic Models. In *Proceedings of the conference on empirical methods in natural language processing*.

33. Roberts, M. E., Stewart, B. M., Tingley, D., Lucas, C., Leder-Luis, J., Gadarian, S. K., … Rand, D. G. (2014). Structural Topic Models for Open-Ended Survey Responses. *American Journal of Political Science*, *58*(4), 1064–1082. https://doi.org/10.1111/ajps.12103

34. Bischof, J. M., & Airoldi, E. M. (2012). Summarizing topical content with word frequency and exclusivity. In *Proceedings of the 29th International Conference on Machine Learning (ICML-12)* (pp. 201-208).

35. Chow, G. C. (1960). Tests of Equality Between Sets of Coefficients in Two Linear Regressions. *Econometrica, 28*(3), 591–605. https://doi.org/10.2307/1910133

36. Zeileis, A., Leisch, F., Hornik, K., & Kleiber, C. (2002). strucchange : An *R* Package for Testing for Structural Change in Linear Regression Models. *Journal of Statistical Software, 7*(2). https://doi.org/10.18637/jss.v007.i02

37. Mather, V. (2019, February 15). A Timeline of Colin Kaepernick vs. the N.F.L. *The New York Times*. Retrieved from https://www.nytimes.com/2019/02/15/sports/nfl-colin-kaepernick-protests-timeline.html

38. Las Vegas shooting: Timeline shows how it unfolded, over one year later - Business Insider. (2019). Retrieved March 1, 2020, from https://www.businessinsider.com/timeline-shows-exactly-how-the-las-vegas-massacre-unfolded-2018-9?r=US&IR=T

39. Behind Berkeley's Semester of Hate - The New York Times. (2017). Retrieved March 1, 2020, from https://www.nytimes.com/2017/08/04/education/edlife/antifa-collective-university-california-berkeley.html

40. Mayer, H. (2018). The alt-right manipulates free-speech rights. We should defend those rights anyway. - The Washington Post. Retrieved February 14, 2020, from https://www.washingtonpost.com/news/made-by-history/wp/2018/08/21/the-alt-right-manipulates-free-speech-rights-we-should-defend-those-rights-anyway/

41. Lewis, H. (2019). How Anti-feminism Is the Gateway to the Far Right. *The Atlantic.* Retrieved February 14, 2020, from https://www.theatlantic.com/international/archive/2019/08/anti-feminism-gateway-far-right/595642/

42. White Nationalist Leads Torch-Bearing Protesters Against Removal of Confederate Statue. (2017). Retrieved March 1, 2020, from https://www.nbcnews.com/news/us-news/white-nationalist-leads-torch-bearing-protesters-against-removal-confederate-statue-n759266

43. Reuters. (2019). Charlottesville: white supremacist gets life sentence for fatal car attack. *The Guardian*. Retrieved from https://www.theguardian.com/us-news/2019/jun/28/charlottesville-james-fields-life-sentence-heather-heyer-car-attack

44. Regner, T. (2020). Crowdfunding a monthly income: an analysis of the membership platform Patreon. *Journal of Cultural Economics*. https://doi.org/10.1007/s10824-020-09381-5